\documentclass[conference]{IEEEtran}
\IEEEoverridecommandlockouts

\usepackage{cite}
\usepackage{amsmath,amssymb,amsfonts}
\usepackage{algorithmic}
\usepackage{siunitx}
\usepackage{graphicx}
\usepackage{float}
\usepackage{textcomp}
\usepackage{svg}
\usepackage{xcolor}
\usepackage[switch]{lineno}
\def\BibTeX{{\rm B\kern-.05em{\sc i\kern-.025em b}\kern-.08em
    T\kern-.1667em\lower.7ex\hbox{E}\kern-.125emX}}
\begin{document}

\title{Levio: Smart Actuated Back-Exoskeleton via Retrofit of a Commercial Passive Device\\

\thanks{This work was supported by the Netherlands Sectorplan Techniek II, the EU Horizon 2024 RIA (Grant No. 101120408; S.W.A.G.), and the Royal Society (IES \textbackslash R2\textbackslash 242059). All authors are affiliated with the Department of Biomechanical Engineering, University of Twente, The Netherlands. $^+$Shared first authorship $^*$Corresponding author:  m.i.mohamedrefai@utwente.nl.}
}

\author{Ioannis Kyriazis$^+$, Mohamed Irfan Refai$^{+,*}$,  Herman van der Kooij}

\maketitle

\begin{abstract}
Passive back exoskeletons show promising benefits for reducing occupational musculoskeletal disorders. However, their assistance cannot be modulated towards the task requirements. Active exoskeletons offer opportunities for pose-specific assistance, but, developing these systems from scratch for research can be time consuming. Therefore, here, we present an accelerated design pipeline that retrofits a Commercial Off-The-Shelf (COTS) passive back exoskeleton, Laevo Flex, with COTS actuation (CubeMars AK10-9) and sensing (IMUs and load-cell) paradigms. Levio was able to provide assistance based on the user's pose. Nine participants performed stoop lifting with the original Laevo and the retrofit Levio system, in transparent and assisted modes. Compared to Laevo, Levio modulated the assistance provided between the bending and lifting phases of the stoop task. The passive assistance offered by Laevo was \(25.2 \pm 4.3~\si{Nm}\) and \(30.6 \pm 9.5~\si{Nm}\) respectively during the bending and lifting phases of the stoop tasks. Levio reduced the resistive torque during bending to \(17.9 \pm 4.6~\si{Nm}\) and offered increased assistance of \(37.7 \pm 6.4~\si{Nm}\) during lifting. Moreover, participants reported higher usability scores and lowest mean global discomfort score while using the Levio when compared to Laevo. These benefits demonstrates retrofitting passive exoskeletons is a viable direction for rapid prototyping of wearable exoskeletons for lifting. 
\end{abstract}

\begin{IEEEkeywords}
Joint impedance, joint stiffness, stance perturbation, rehabilitation, system identification.
\end{IEEEkeywords}

\section{Introduction}

Musculoskeletal disorders (MSDs) chronically affect about 120 million people in the EU~\cite{Crawford2021}. Occupational MSDs, driven largely by repetitive load handling and trunk bending, account for close to 50\% of workplace absences~\cite{Crawford2021, Bevan2015}, with low-back injury a major consequence~\cite{heneweer2011physical,wai2010lifting}. Automation can remove some manual handling, but tasks that demand variability and adaptability require human dexterity, making wearable assistance a promising complement~\cite{Toxiri2019}.

Passive back-support exoskeletons store and release energy in viscoelastic elements and reduce low-back muscle activity by roughly 10--40\% during lifting~\cite{refai2024benchmarking, Schmalz2022, Schnieders2023, laevo2022torque}. As, the magnitude and timing of assistance are fixed by the device's geometry, impedance properties, and hysteresis~\cite{koopman2020passive, koopman2020spexor, refai2024benchmarking, Ali2021}, they cannot adapt to the user's intent, movement phase, or carried load. Active exoskeletons address this limitation with powered, programmable assistance. Studies show that active assistance can reduce low-back muscle activity upto 30\% \cite{moya2025neuromechanical, chung2024active, Lei2025} and low-back compressive forces by 5--28\% during lifting~\cite{Koopman2019, chung2024active, moya2025neuromechanical, Hu2026}. As commercial active back exoskeletons are generally closed to research interfacing and control, research studies rely on purpose-built prototypes~\cite{chung2024active, Cullen2026, moya2025neuromechanical}. However, developing these prototypes requires stable exoskeleton-user interfaces that enable nominal force transfer from the actuation system. This increases the iteration cycle time, fabrication costs, and still result in limited adaptability to different body sizes \cite{moya2025neuromechanical, Brouwer2024}.

Retrofitting commercial passive exoskeletons and orthoses addresses each of these barriers. The exoskeleton-user interfaces of passive systems are designed to fit a range of body dimensions and have been validated in user populations~\cite{Albouy2025, Kim2025, NationalSafetyCouncil2025, Jakobsen2025}, which removes the need to redesign the interface and shortens iteration cycles and cost. The passive assistance of commercially available exoskeletons is also well documented~\cite{laevo2022torque, refai2024benchmarking}, which provides a direct baseline for quantifying the added value of active assistance. Additionally, a lightweight retrofit could be offered as an add-on module for an already commercial device, broadening the device's usability. Retrofit approaches have been demonstrated mainly for the lower limb, using commercial braces with commercial or custom actuation units~\cite{Nesler2022, Payra2021, Pergolini2025, Scheidl2026}.

Retrofitting is not trivial, however. The frame and interfaces of passive systems were designed around their integrated viscoelastic elements, so adding actuation and sensing requires careful mechanical coupling with these elements. In our previous work, we integrated the Laevo Flex chest harness into an active back-exoskeleton design as it offered a comfortable interface across genders and body dimensions~\cite{refai2024benchmarking, Hu2026, Brouwer2024}. Existing retrofit studies address the knee and hip for gait, sit-to-stand, or squat lifting~\cite{Payra2021, Nesler2022, Pergolini2025, Scheidl2026}, and typically compare against no device rather than the original passive system. To our knowledge, there is therefore no documented design methodology for actively retrofitting a passive back-support exoskeleton, nor evidence of how such a retrofit performs against its passive baseline during lifting.

Therefore, this paper presents Levio, an active retrofit of the passive Laevo FLEX V3.0. As the elastic properties of the Laevo exoskeleton is well documented ~\cite{laevo2022torque}, it offers a direct benchmark to build the active system. We integrated COTS actuators, sensing, and electronics, and implemented a phase-dependent controller. Our main research question was \emph{what is the biomechanical impact of a retrofitted back exoskeleton with COTS actuation?} Specifically, we (i) identify the actuator and drivetrain characteristics required to match or improve the assistance of the passive device, (ii) develop the mechanical modifications and electronics required to integrate the actuation, and (iii) compare the retrofit with the passive system in performance and comfort.


\section{Requirements Analysis}
In our earlier benchmarking study, we found that the rigid passive back-exoskeleton Laevo FLEX V3.0 provided the most assistance while also being the most comfortable \cite{refai2024benchmarking}. Moreover, the assistance profiles of the system are well documented \cite{laevo2022torque, Schnieders2023, refai2024benchmarking, koopman2020passive}, and we had access to an exoskeleton for modification. Therefore, we selected this as our retrofit platform. 

Laevo offers nonlinear support during sagittal rotation \cite{laevo2022torque, refai2024benchmarking}. The device is shipped with five springs which provide bilateral support ranging from approximately 20 to 60~Nm, with the respective highest support occurring during the bending phase \cite{laevo2022torque}. Laevo has additional passive alignment joints close to the hips to accommodate stance-width changes and out-of-plane thigh motion without generating assistance in those directions. Force-transfer bars route the assistance between the trunk and thighs. A rear ball-and-socket joint placed posteriorly on the trunk provides additional rotational freedom, allowing the force-transfer bars to reorient itself relative to the vest and maintain an approximately perpendicular force, even during asymmetric movement.

The retrofit requirements were derived from the passive Laevo characteristics, the constraints of its existing structure, and the additional functions required for active assistance. A V-model approach was followed to link system requirements, subsystem development, and prototype verification. To retain the passive system's functionality, the retrofit aimed to preserve the existing torso and thigh interfaces, sagittal bending range, and the additional passive degrees of freedom accommodating out-of-plane movements. Moreover, the added components should be integrated into the mechanical frame, and ensure back-drivability of the drivetrain for free user movement during unassisted modes.

We selected a maximum target actuation of \SI{80}{Nm}, with \SI{40}{Nm} per side, corresponding to the strongest support available via Laevo (\SI{60}{Nm}), with room for additional assistance. The drivetrain should permit \(0\text{--}120^\circ\) movement range and a design speed of 120°/s\cite{marrasTrunkForceDevelopment,granataSpinalLoadingLoweringLifting}. This requires designing new load-bearing components for \SI{40}{Nm} per side with a target factor of safety of two.

Active operation required repeatable state-dependent torque profiles, angle and angular-velocity sensing, calibrated rear load-path force measurement, actuator communication, and data logging. The system also had to support active assistance, a transparent mode with the motors enabled and commanded to zero torque, and manual backdrivability when unpowered. Configurable torque limits, protected power distribution, fault handling, and independent hardware emergency power removal were also required.

\section{Device Design}

The Levio system retained the frame and user interfaces of the Laevo system. The added actuators, sensing elements, electronics, and power components are described further in the following sections.

\subsection{Actuation System}

To select the appropriate drivetrain system, several architectures were considered. These systems were screened against torque capability, form factor, manual backdrivability, and suitability for testing different controllers. Direct drive actuators were too large for the required torque, while linear and fluid-power concepts conflicted with the existing Laevo frame and portability. Cable transmission and high-ratio gearing introduced additional compliance, friction, or backdrive resistance. Low-ratio quasi-direct-drive (QDD) actuation was therefore selected as the best overall compromise.

Following drivetrain selection, candidate QDD actuators were screened for torque--speed capability, electrical feasibility, peak current demand, and thermal loading based on representative bending and lifting cycles \cite{laevo2022torque}. The final shortlist consisted of the RobStride-03 (RobStride, Beijing, China) \cite{robstride03}, CubeMars AKE80-8 KV30, and CubeMars AK10-9 V3.0 KV60 (CubeMars, Nanchang, China) \cite{cubemarsgeneral}.

The RobStride-03 satisfied the calculated requirements but was unavailable within the project schedule. The AKE80-8 could provide a torque of \SI{30 }{Nm} but could not deliver it at the design speed of 120°/s within the available voltage limit. Finally, the AK10-9 satisfied the \SI{40 }{Nm} component-level target while providing suitable dimensions, availability, encoder configuration, and compatibility with the external controller electronics. It was therefore selected for the retrofit.

\subsection{Mechanical Components for Retrofitting}

\begin{figure}[!t]
\centering
\includegraphics[width=0.5\textwidth]{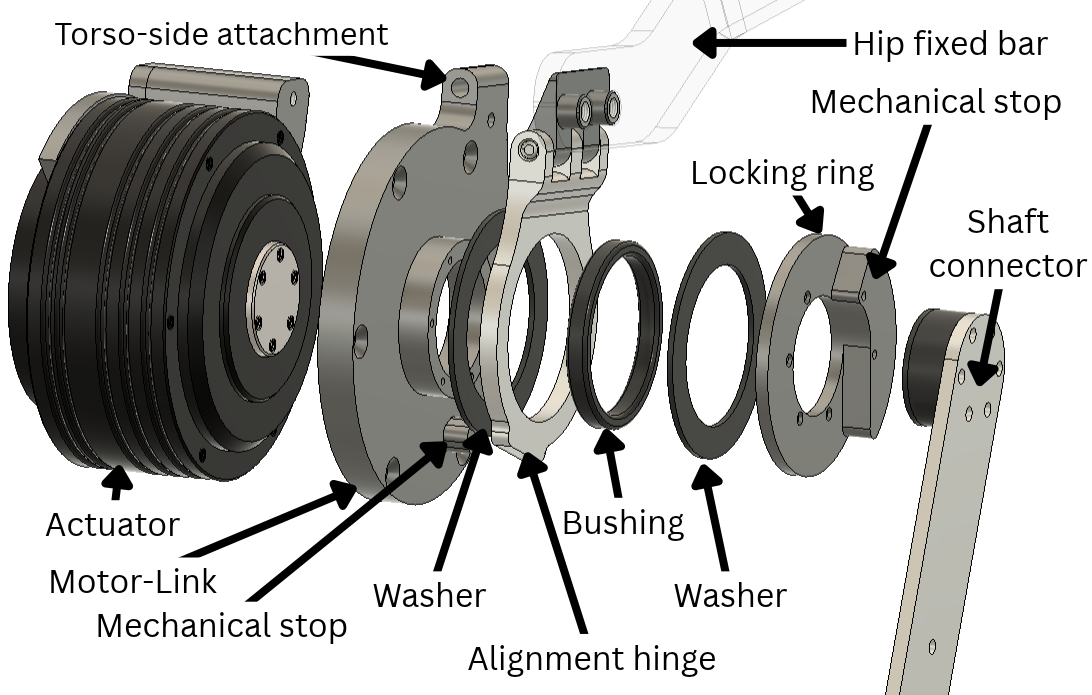}
\caption{Exploded-view of the mechanical frame for integrating the actuation system. The alignment hinge allowed preservation of passive degrees of freedom. }
\label{fig:mechanical-assembly-sequence}
\end{figure}

The mechanical retrofit was designed to integrate the selected actuators into the Laevo frame while retaining its torso, vest, and thigh interfaces. The design had to preserve the required bending range and stance-width adjustment, fit within the existing frame, withstand the 40 Nm per-side design load, and avoid unacceptable binding during movement.

Figure~\ref{fig:mechanical-assembly-sequence} shows an exploded-view of the mechanical components for integration of the actuation system. As, the original passive structure already transfers an extension moment between the torso and thigh interfaces, the retrofit preserves this load path. The Laevo gas-spring is replaced with CAM mechanism with two CubeMars AK10-9 actuators, one for each side of Levio. The added components connect each actuator between the torso-side and thigh-side structures so that the commanded torque produces an opposing torque pair between them.

The mechanical assembly includes a separate alignment hinge. Each side provides two distinct rotational degrees of freedom. The actuated degree of freedom is located between the torso-side structure and the thigh-side shaft, generating the sagittal-plane assistance moment. The passive alignment hinge allows the complete actuator and shaft assembly to move laterally to accommodate hip abduction and adduction, keeping the thigh-side shaft approximately parallel to the leg during changes in stance width. The hinge, however, does not generate assistance.

    \label{fig:mechanical-retrofit-side-view}

Assistance is transferred through two concentric load paths. A tubular motor link, rigidly connected to the actuator case, carries torque through the force-transfer bars to the torso-side structure (the vest and back interface). The extended output shaft passes through this tube and connects via the shaft connector to the thigh-side shaft, carrying the opposing torque to the thigh interfaces. The relative torque between the two paths produces the extension assistance.

\begin{figure}[!t]
    \centering
    \includegraphics[width=0.45\textwidth]{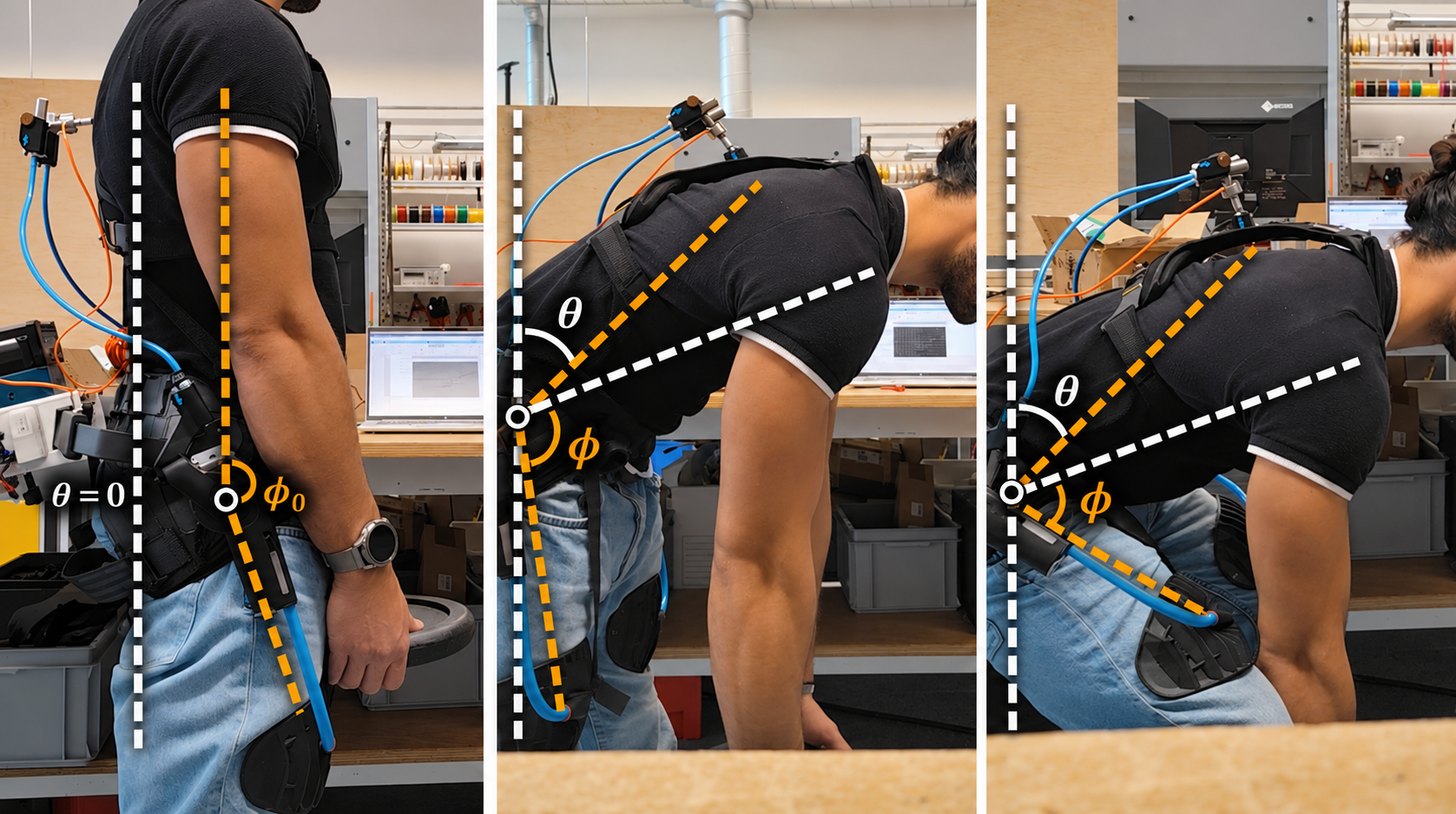}
    \caption{Measured torso--pelvis angle $\theta$ and effective exoskeleton angle $\phi$ in upright (left), stoop (middle), and squat (right) postures.}
    \label{fig:measured-effective-angle-comparison}
\end{figure}

\subsection{Sensing and Communication}
Inertial and force sensors were integrated for kinematic and interaction measurements required for control and verification.

A load-cell was integrated at the rear between the vest and the joint that connect the passive rods. This allowed us to measure the assistance transmitted through the torso interface. An equivalent assistance
torque can be inferred when the corresponding effective moment arm is defined. 

Further, two ICM-42688-P inertial measurement units (IMUs) (TDK InvenSense, San Jose, CA, USA) were mounted on the torso and pelvis. Figure~\ref{fig:measured-effective-angle-comparison} illustrates two angle definitions, \(\theta\) and \(\phi\), where \(\theta\) is defined as the relative torso--pelvis angle,  $\theta_{torso} - \theta_{pelvis}$, measured using the two IMUs \cite{brouwer2021trunk}. However, the effective exoskeleton angle \(\phi\) is denoted by $\theta_{torso} - \theta_{thigh}$ \cite{laevo2022torque}. As no thigh IMU or exoskeleton-joint encoder was included in the implemented prototype, \(\phi\) was not measured directly . Figure~\ref{fig:measured-effective-angle-comparison} demonstrates that for stoop tasks, \(\theta\) and \(\phi\) follow a similar trend, whereas this would not be the case for squat tasks. This motivated our choice for measuring stoop tasks. The corresponding relative angular velocity could be used to distinguish bending, low-velocity holding, and lifting states.

A Raspberry Pi 4b 2Gb (RPI) was used as the embedded microcontroller. The original integrated AK10-9 motor controllers were replaced by MAB Robotics MD80 controllers (MAB Robotics, Poznań, Poland) \cite{mab_md_series}. The RPI communicated with both MD80 motor controllers through the CAN FD network using the CANdle interface. The controllers receive torque commands and return actuator position, velocity, estimated motor torque, temperature, supply voltage, and fault information where available. These signals are used for monitoring, logging, and fault handling. 

\subsection{Control Design}
\begin{figure}[!t]
    \centering
    \includegraphics[width=0.5\textwidth]
    {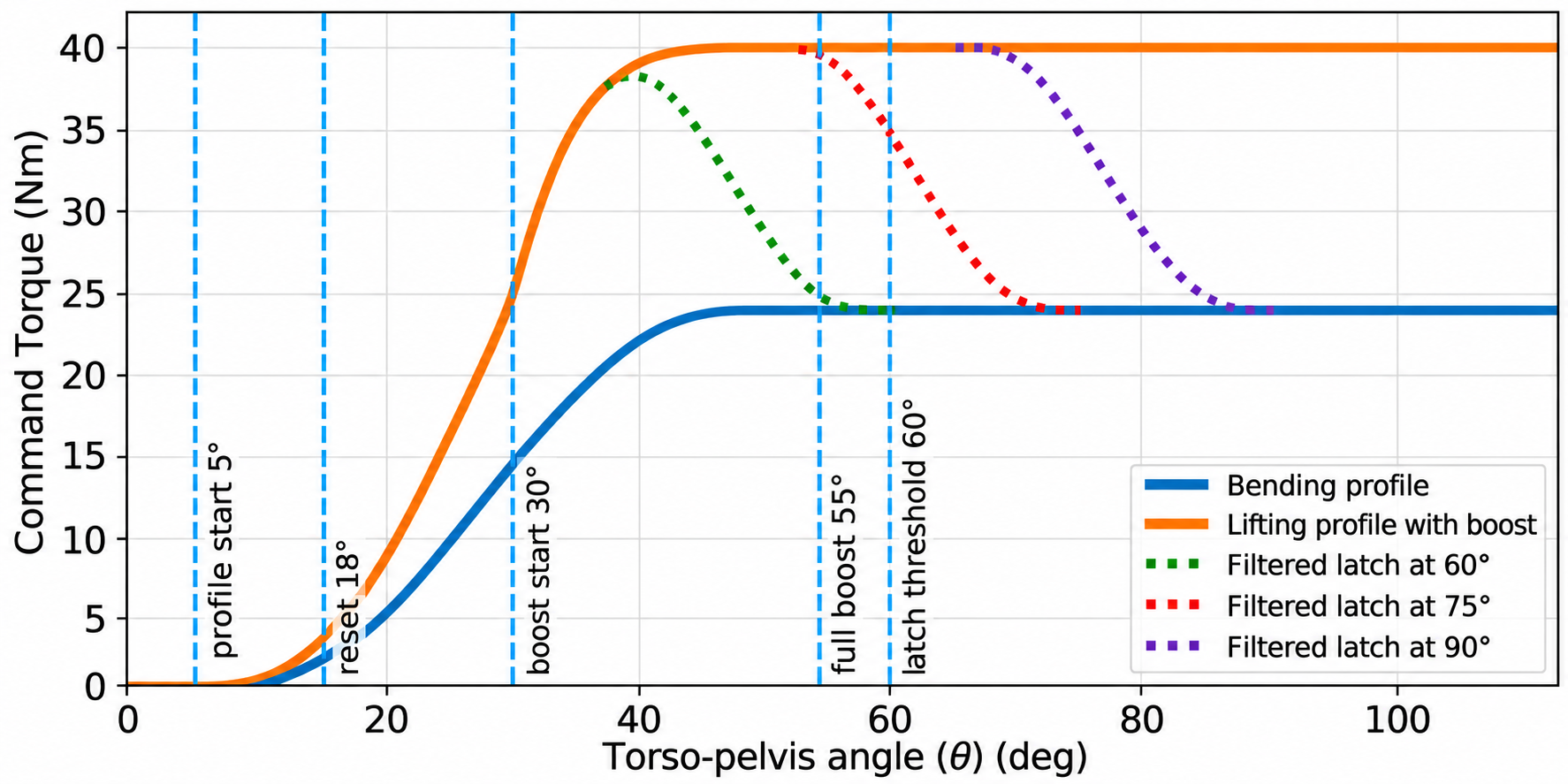}
    \caption{Target nominal bending and lifting assistance command torques. }
    \label{fig:assistance-profiles}
\end{figure}

The assistive controller implemented a state-dependent kinematic torque command as seen in Figure \ref{fig:assistance-profiles}. The command used a smooth posture-dependent shape as a starting point, with reduced assistance during bending and higher assistance during lifting \cite{lazzaroni2020acceleration,lazzaroni2022efficacy,chung2024active}.

\begin{figure*}[!ht]
\centering
\includegraphics[width=0.7\textwidth]{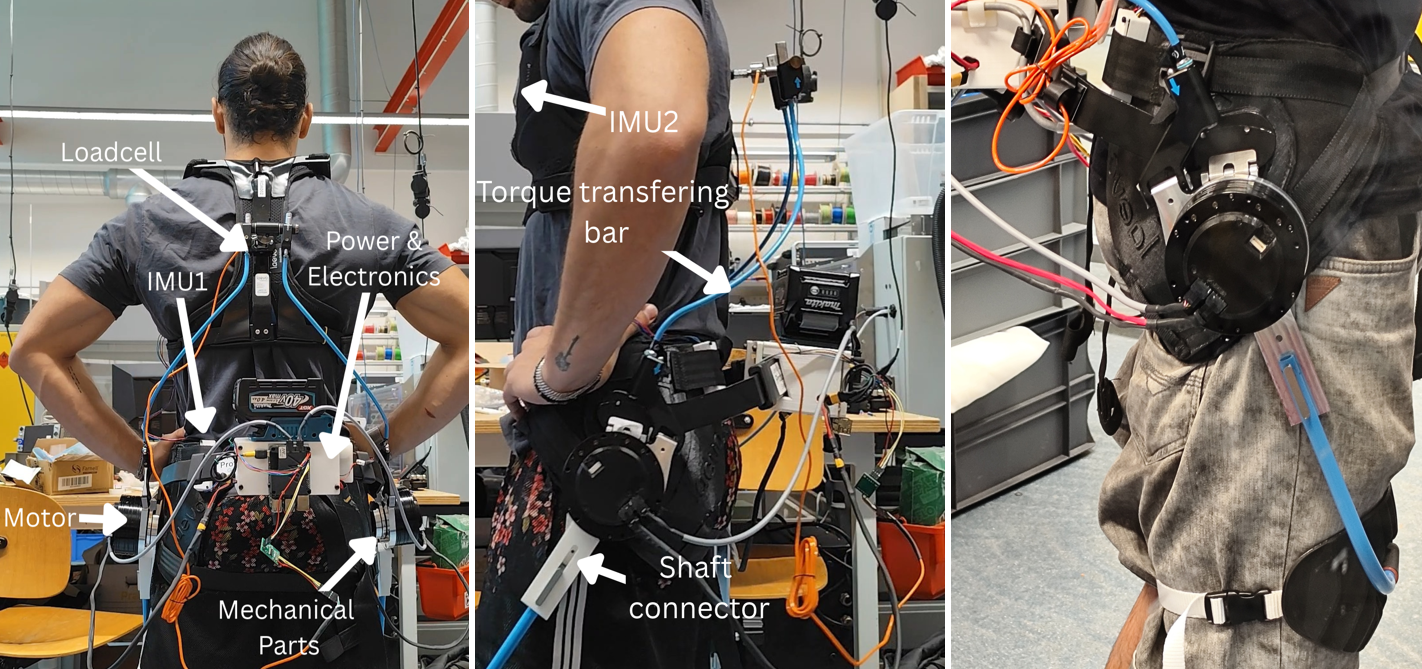}
\caption{Back and side profiles of the Levio system with the Laevo frame and integrated active components. The actuator is positioned between the torso-side force-transfer bars and the thigh-side structure, while the separate alignment hinge accommodates out-of-plane leg motion. The first two panels include 3D-printed prototype parts, whereas the rightmost panel shows the final version with machined parts.}
\label{fig:levio-system-overview}
\end{figure*}

The RPI executed the high-level controller at approximately 50 Hz, the highest stable rate achieved with the current software implementation. Higher rates may be possible through software optimisation. The controller computes torque commands from two predefined assistance profiles for bending and lifting. It follows the bending profile until angle and angular-velocity conditions trigger the lifting state, which remains latched until returning toward upright. Transitions are smoothed through time-dependent blending, filtering, and torque rate limiting.

The measured torso--pelvis angle \(\theta\) is normalized to \(x\in[0,1]\) between \(5^\circ\) and \(50^\circ\), allowing the smootherstep function to smoothly scale the base assistance profile from zero to its maximum:
\begin{equation}
    s(x)=6x^{5}-15x^{4}+10x^{3},
    \qquad
    x=
    \operatorname{clip}
    \left(
    \frac{\theta-\SI{5}{\degree}}
    {\SI{45}{\degree}},
    0,1
    \right),
    \label{eq:smootherstep-profile}
\end{equation}

with the nominal full and bending profiles defined as

\begin{equation}
    \tau_{\mathrm{full}}=T_{\max}s(x),\quad
    \tau_{\mathrm{bend}}=0.60T_{\max}s(x),\quad
    T_{\max}=40\,\mathrm{Nm}.
    \label{eq:assistance-profiles}
\end{equation}

The bending profile uses \(60\%\) of the nominal per-side assistance, with \(T_{\max}\) representing the controller design value of \SI{40}{Nm} per side.

The transition from bending to lifting is triggered when
\[
    \theta\geq\SI{60}{\degree},
    \qquad
    \lvert\dot{\theta}\rvert
    \leq
    \SI{4}{\degree\per\second}.
\]
Once the latch is activated, a time-dependent lifting weight \(w\) transitions gradually from the bending toward the lifting profile, with, target assistance defined as 

\begin{equation}
    \tau_{\mathrm{target}}
    =
    (1-w)\tau_{\mathrm{bend}}
    +
    w\tau_{\mathrm{lift}},
    \qquad
    0\leq w\leq1.
    \label{eq:controller-profile-blend}
\end{equation}

The lifting state remains latched during the return movement and resets to the bending state when the measured angle falls below \SI{18}{\degree}. Moreover, the lifting profile includes an additional angle-dependent boost toward \(T_{\max}\).
The boost acts on top of the full smootherstep lifting profile as the angle decreases from approximately \SI{55}{\degree} to \SI{30}{\degree}. In this range, it compensates part of the decrease of the underlying smootherstep, so that lifting assistance falls more gradually and remains higher over a larger portion of the return movement. Above this range the smootherstep is already at, or close to, \(T_{\max}\), while below approximately \SI{30}{\degree} the boost is removed. This approach was used to extend the high-assistance region without redesigning the underlying lifting
profile.

The blended torque target is passed through a low-pass filter and a torque slew-rate limiter before being sent to the motors. The low-pass filter smooths the command by attenuating rapid variations, such as those from measurement noise or discrete controller-state updates. The slew-rate limiter separately caps the rate of change of the commanded torque. It limits how fast assistance can rise, while a higher permitted rate of decrease lets assistance be removed quickly when required. Together, these steps give a smoother command and limit abrupt actuator loading, at the cost of added response delay and potentially reduced peak assistance during rapid movements.

\subsection{Power Electronics}

The electronics and power subsystem supplies the actuators and control electronics through separate protected branches and provides independent hardware emergency power removal. The main design challenge was combining the high short-duration current required by the actuators with the stable low-voltage supply required by the control electronics.

We utilized a removable Makita battery operating between \SI{36}{V} and \SI{42}{V}. A DC contactor switches the battery output before reaching the downstream distribution point. The battery was selected for its burst-current capability, availability, and removable format.

A contactor coil is supplied by a dedicated \SI{12}{V} DC/DC converter connected upstream of the contactor. A normally closed emergency-stop switch interrupts the coil supply. Pressing the emergency stop de-energises the coil, opens the contactor, and disconnects the downstream power branches.

After the contactor, power is divided into two separately fused motor-controller branches and one low-voltage logic branch. The motor branches supply battery voltage to the left and right MD80 controllers and are each protected by a \SI{30}{A} fuse. The logic branch feeds a DC/DC converter that provides \SI{5}{V} to the Raspberry Pi, CAN interface, and sensing electronics. Fuses before and after the converter protect its input and the downstream \SI{5}{V} wiring.

Safety was implemented through complementary mechanical, electrical, and software measures, providing protection without relying exclusively on the high-level controller.

For brief periods of invalid IMU data, the most recent valid measurement is retained. Persistent sensor faults or critical communication faults trigger a zero-torque command and termination of active control. The independent hardware emergency-stop button allows immediate removal of actuator power.

\subsection{Integrated Levio System}

Figure~\ref{fig:levio-system-overview} shows the Levio system that retains the frame and user interfaces of the Laevo system. The completed Levio prototype had a total mass of approximately 7 kg, whereas the original Laevo system weighed 4.3 kg.

\begin{figure}[!t]
\centering
\includegraphics[trim = 0.4cm 0.4cm 0.9cm 0.3cm,clip,width=0.45\textwidth]{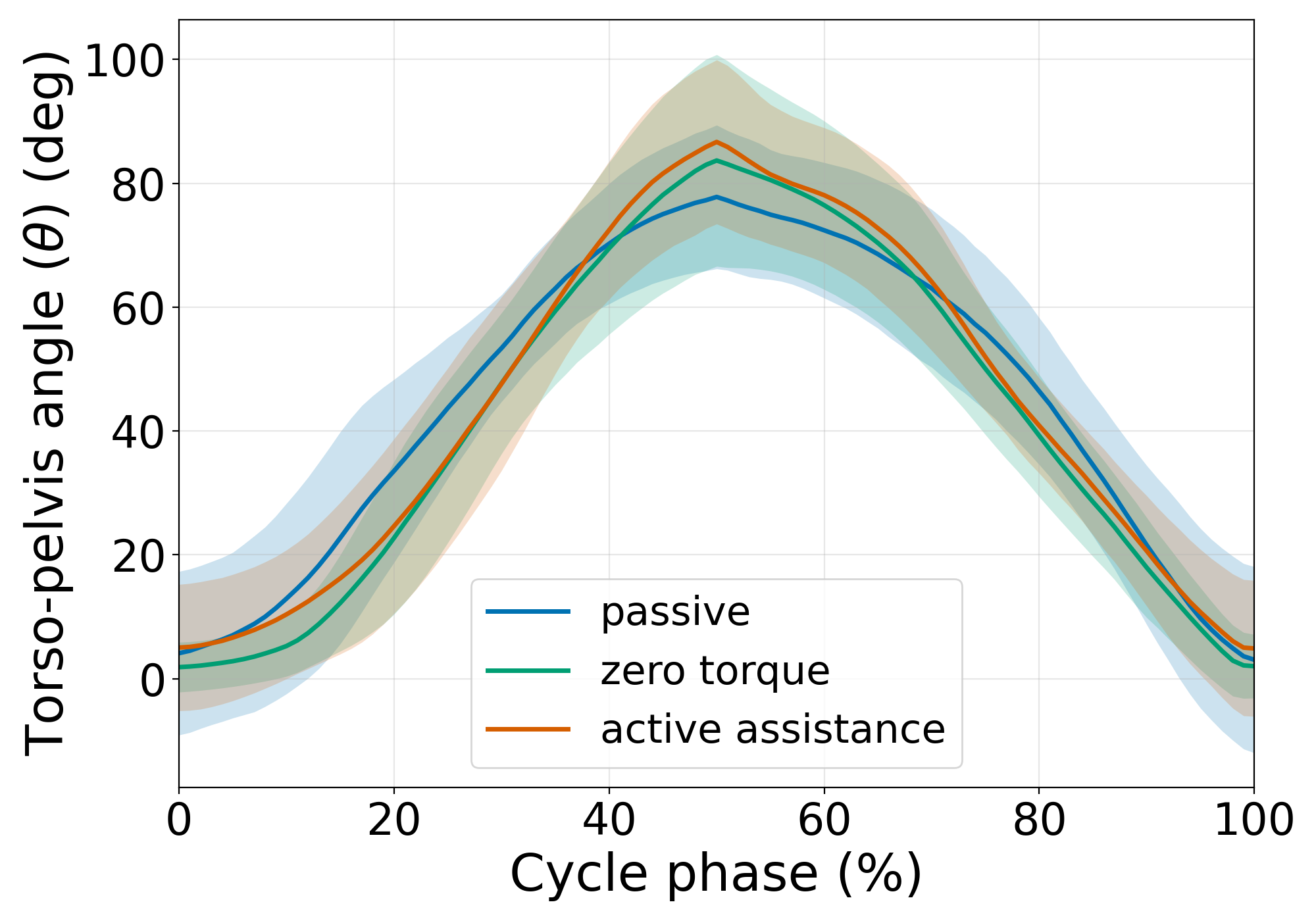}
\caption{Cycle normalized average torso-pelvis angle (\(\theta\)) measured across the three lifting conditions. The shaded region shows the standard deviation across participants.}
\label{fig:angle_across_conditions}
\end{figure}

\section{Participant Testing}

Before participant testing, integrated verification included mechanical range-of-motion checks, bilateral actuator operation, sensor and communication checks, and emergency-stop functionality.

Following fitting and familiarisation, participants performed at least five repetitions of stoop lifting under three conditions: i) \textit{passive} (wearing the passive Laevo), ii) \textit{zero torque} (Levio in zero-torque mode), and iii) \textit{active assistance} (Levio with active assistance mode). For all conditions, participants lifted with a \SI{10}{kg} disc. The movement timing was standardised using a metronome. The experiments were conducted with the approval of the University of Twente Ethical Committee 261523, and participant provided written informed consent.

\subsection{Data collection}
The participant evaluation focused on comapring the mechanical and subjective outcomes between the passive reference and the active operating modes. For all conditions, IMU and load-cell readings were measured. Participants were asked to pause momentarily at the upright and maximally flexed positions, for easier cycle segmentation. Lifting cycles were segmented using the \(\theta\) trajectory and separated into bending and lifting phases. Incomplete movements and local fluctuations that did not represent complete cycles were removed from further analysis. 

We measured the assistive force measured by the load-cell at the rear, the commanded torque, and \(\theta\) across the conditions. The equivalent rear-interface assistance torque (\(\tau_{estimated}\)) was estimated by multiplying the measured load-cell force by the effective moment arm. The moment arm was measured as the distance from the attachment of the back hinge to the joint rotation of the actuators, as 0.45 m. The \(\tau_{estimated}\) quantifies the assistance transmitted to the user through the rear vest attachment, while the commanded torque and controller state were used to relate changes in this force to the timing and magnitude of the active assistance.

For each participant, we also estimated the difference in \(\tau_{estimated}\) provided by the \textit{passive} and \textit{active assistance} conditions. These were estimated from the participant-level mean curves before group averaging. The curves were aligned by relative torso--pelvis angle, between \(18^\circ\)--\(70^\circ\) interval. In the \textit{zero-torque} command, residual interaction was quantified by measuring phase-specific root-mean-square (RMS) and peak absolute rear-interface torque.

Participants were also requested to fill in objective questionnaires, the standard 10-item System Usability Scale (SUS) \cite{brooke1996sus}, global discomfort and body-region discomfort scores. Discomfort was evaluated using a custom 0--10 visual analogue scale (VAS) questionnaire \cite{refai2024benchmarking, huskisson1974pain} These scores were summarised descriptively by condition. No inferential statistics were applied because the participant evaluation was used for engineering verification rather than hypothesis testing. 

\begin{figure}[!t]
\centering
\includegraphics[trim = 0.4cm 0.4cm 0.4cm 0.3cm,clip,width=0.45\textwidth]{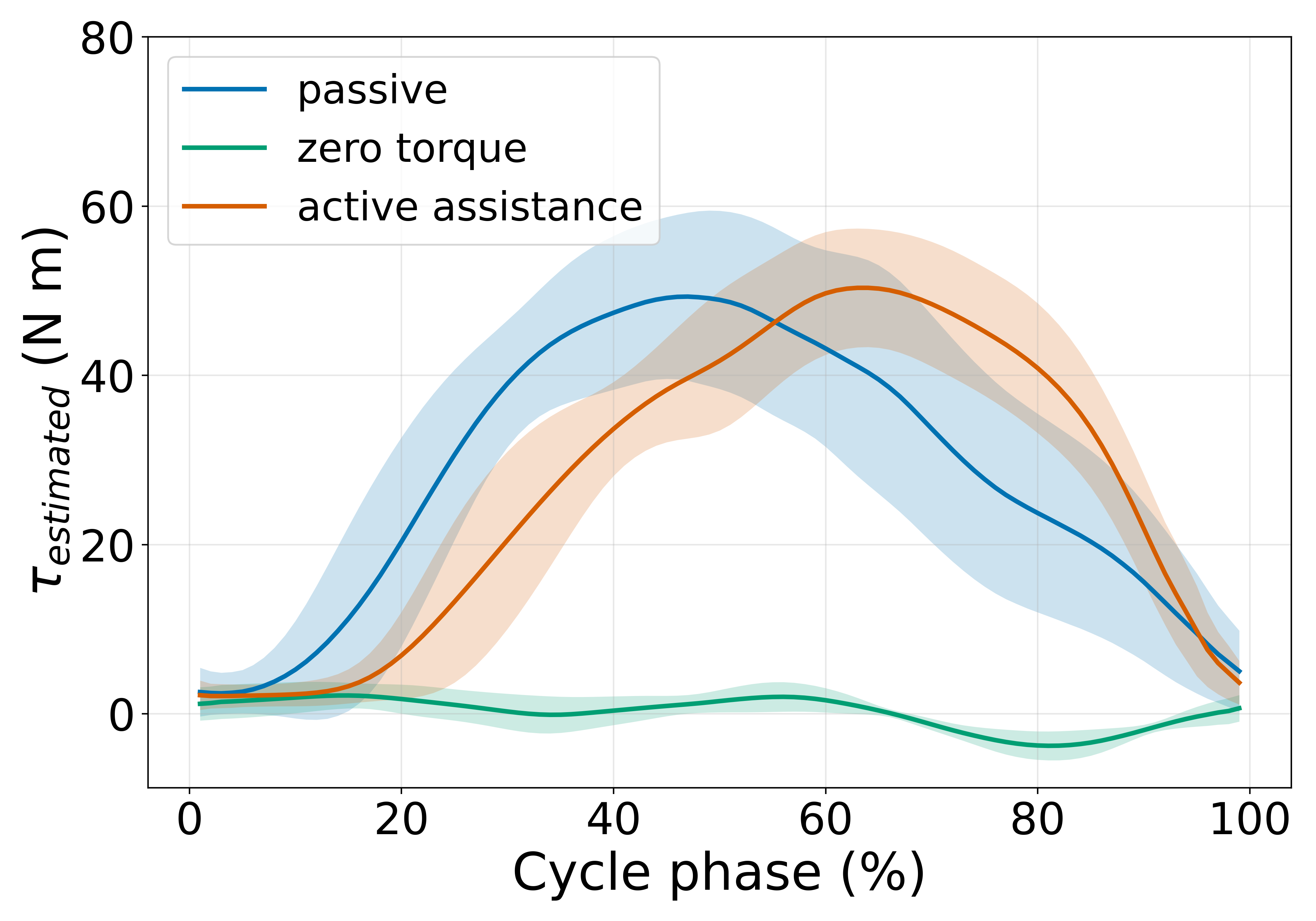}
\caption{Cycle normalized average assistive torques \(\tau_{estimated}\) across conditions. The shaded region shows the standard deviation across participants.}
\label{fig:tau_across_conditions}
\end{figure}

\section{Results}
During integrated mechanical verification, we observed deformation of the thigh torque-transfer connectors when the assistive torques reached 40 Nm on either side. This indicated that the preliminary FEA had overestimated the structural capacity of the redesigned connectors. To proceed with participant testing, we reinforced the connector with an additional parallel bar, and limited the maximum commanded torque to 25 Nm per side. Within the reduced operating limit, the prototype completed the tested movements without further visible deformation or binding. Bilateral actuation, sensing, communication, logging, and hardware emergency power removal operated as intended during integrated testing.

Nine participants (seven male and two female) were recruited for the experimental protocol, with ages ranging from 25 to 31 years, heights from 171 to 191 cm, and body masses from 65 to 91 kg. Due to data collection issues, only seven participants were included in the objective assistance-torque analysis, whereas data from all nine participants were used for the subjective questionnaire results.

\begin{figure}[!t]
\centering
\includegraphics[trim = 0.4cm 0.3cm 0.4cm 0.4cm,clip,width=0.5\textwidth]{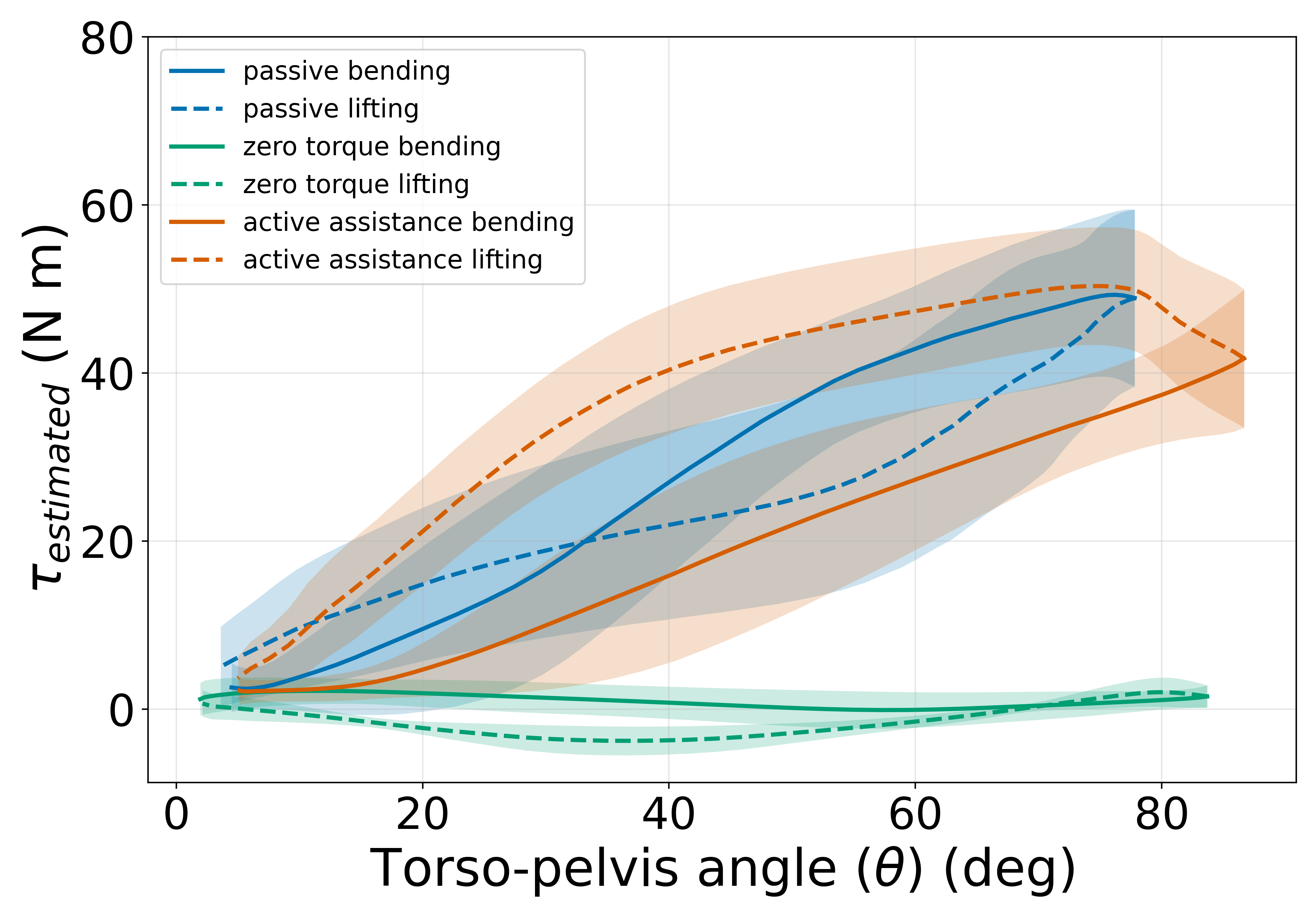}
\caption{Work-loops demonstrating the relation between \(\tau_{estimated}\) and \(\theta\) across the three conditions. Solid line denotes the bending phase, whereas the dotted line denotes the lifting phase of the respective conditions.}
\label{fig:group-loadcell-force-angle}
\end{figure}

We found that, overall, participants took approximately \SI{5}{s} each, for bending and lifting during the \textit{passive} condition. However, this was shorter, approximately \SI{4}{s}, for the \textit{zero torque} and \textit{active assistance} conditions. The shorter phases helped reduce participant fatigue while getting accustomed to the Levio system. We also found no noticeable differences between the \SI{4}{s} and \SI{5}{s} cycle times during preliminary analyses. 

Fig. \ref{fig:angle_across_conditions} shows the average torso--pelvis angle \(\theta\) across conditions. We see no noticeable difference in the kinematics across the three conditions for the stoop lifting task. Fig. \ref{fig:tau_across_conditions} demonstrates the corresponding \(\tau_{estimated}\) measured during the three conditions. We see that while wearing the Laevo in \textit{passive} condition, the peak assistance is provided earlier, and occurs during the bending phase of the cycle. However, while wearing the Levio in \textit{active assistance} condition, peak assistance is provided when the participant starts to lift the load and returns to upright stance. Moreover, the \textit{zero torque} condition offers no assistance as expected, with residual interaction RMS values of \(1.9 \pm 1.0\)~Nm during bending and \(2.7 \pm 1.2\) Nm during lifting.

Fig. ~\ref{fig:group-loadcell-force-angle} demonstrates the work-loops between \(\tau_{estimated}\) and \(\theta\) across the three conditions. Solid lines denote the bending phase, whereas the dotted line denotes the lifting phase of the respective conditions. During the \textit{passive} condition, mean \(\tau_{estimated}\) during bending was \(25.2 \pm 4.3\) Nm and \(17.9 \pm 4.6\) Nm for \textit{active assistance}. During lifting, the values were \(30.6 \pm 9.5\) Nm and \(37.7 \pm 6.4\) Nm, for the \textit{passive} and \textit{active assistance} conditions, respectively. During the \textit{passive} condition, we can distinctively see hysteresis between the bending and lifting phases due to the energy stored in the elastic springs. We notice that the initial support (until ~30$^{\circ}$) is lower during the lifting phase compared to the bending phase. During the \textit{active assistance} condition, we see the energy injected into the system, with a larger assistive torque during the lifting phase as compared to the initial bending.





\begin{figure}[!t]
\centering
\includegraphics[trim = 0.4cm 0.3cm 0.4cm 0.4cm,clip,width=0.45\textwidth]{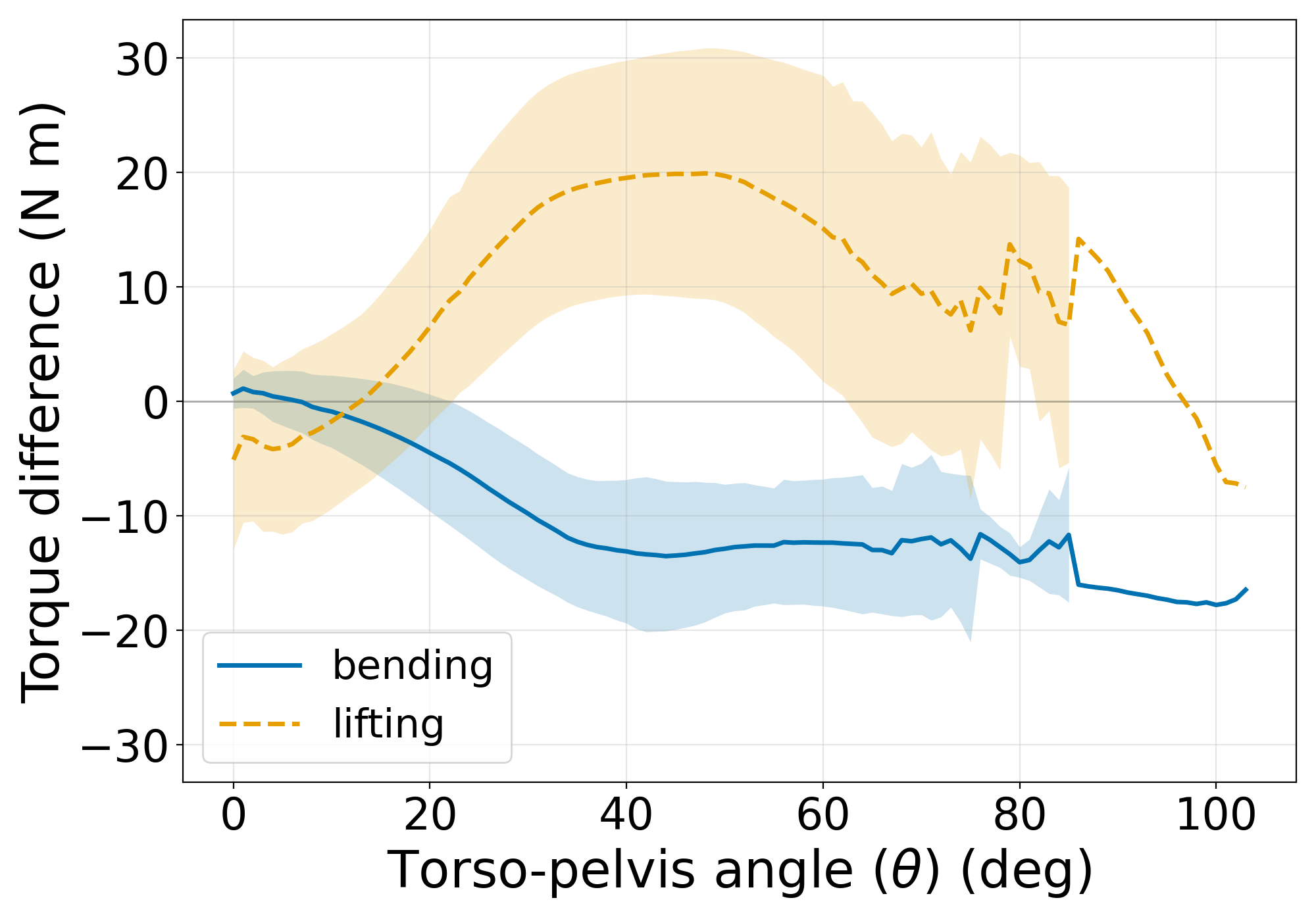}
\caption{Average difference in assistance between the conditions measured as $\textit{active assistance} - \textit{passive}$. Positive torques demonstrate higher assistance during the \textit{active assistance} condition. The shaded region shows the standard deviation across participants.}
\label{fig:active-passive-difference}
\end{figure}

The modulation of assistance by the \textit{active assistance} condition can be visualized in Fig.~\ref{fig:active-passive-difference}. The figure demonstrates the average difference in the assistance provided across conditions, measured as $ \textit{active assistance} - \textit{passive}$. The differences over the \(18^\circ\)--\(70^\circ\) range were \(-11.95 \pm 4.35\) Nm during bending and \(14.76 \pm 10.31\) Nm during lifting. Across this interval, we found that \(\tau_{estimated}\) was lower for the \textit{active assistance} compared to \textit{passive} condition for \(98.4 \pm 3.0\%\) of the bending range, whereas during lifting, \textit{active assistance} was higher for \(88.9 \pm 20.0\%\) of the range. 

Subjective scores, which include the SUS, global and body-region discomfort scores, are shown in tables~\ref{tab:sus-results}, \ref{tab:global-vas-results}, and  \ref{tab:body-region-vas-results}, respectively. The SUS results show that \textit{active assistance} produced the highest mean SUS score, followed by \textit{passive} and \textit{zero torque conditions}. The \textit{zero torque} condition showed the largest between-participant spread. However, perceived usability varied between participants. Across the discomfort measures reported, \textit{active assistance} was scored lowest, for both the global discomfort score and for each recorded body region. The largest reduction was observed at the lower back.

\begin{table}[!t]
\centering
\caption{System Usability Scale results for stoop ($n=9$).}
\label{tab:sus-results}
\small
\setlength{\tabcolsep}{3pt}
\begin{tabular}{lcc}
\hline
\textbf{Condition} & \textbf{Mean $\pm$ SD} & \textbf{Median [range]} \\
\hline
Passive & $76.9 \pm 13.2$ & $72.5$ [62.5--97.5] \\
Zero torque & $65.3 \pm 26.6$ & $72.5$ [15.0--95.0] \\
Active assistance & $81.7 \pm 12.6$ & $77.5$ [60.0--100.0] \\
\hline
\end{tabular}
\end{table}

\begin{table}[!t]
\centering
\caption{Global discomfort VAS results for stoop ($n=9$).}
\label{tab:global-vas-results}
\small
\setlength{\tabcolsep}{3pt}
\begin{tabular}{lcc}
\hline
\textbf{Condition} & \textbf{Mean $\pm$ SD} & \textbf{Median [range]} \\
\hline
Passive & $3.6 \pm 1.0$ & $3.0$ [2.0--5.0] \\
Zero torque & $3.8 \pm 1.9$ & $4.0$ [2.0--7.0] \\
Active assistance & $2.6 \pm 1.6$ & $2.0$ [1.0--6.0] \\
\hline
\end{tabular}
\end{table}

\section{Discussions}
In this study, we designed Levio by retrofitting COTS actuation and sensing onto the commercial Laevo Flex exoskeleton, and compared its biomechanical impact against the passive system during stoop lifting. \textit{Active assistance} did not alter lifting kinematics (Fig.~\ref{fig:angle_across_conditions}), but it modulated assistance timing by task phase (bending or lifting), shifting the peak assistance later (Fig.~\ref{fig:tau_across_conditions}). The passive system support peaks just before the user is fully bent, whereas Levio assistance peaks as the user opposes gravity to return upright, which is where support is most beneficial. Note that the rear-force measurements reflect interface assistance, not direct estimates of lumbar unloading.

This shift came from a deliberately simple state-dependent kinematic torque command \cite{chung2024active}. It uses IMU data and demonstrates that phase and movement direction dependent profile with tuned magnitude and timing can meaningfully change user--device interaction. This provides a controlled baseline before more complex architectures, such as CEINMS-RT or lift-timing models based on force or EMG sensing can be implemented~\cite{Koopman2019, lazzaroni2020acceleration, Sartori2025}.

Fig.~\ref{fig:group-loadcell-force-angle} shows the load-cell-estimated assistance ($\tau_{estimated}$) against torso--pelvis angle ($\theta$). The passive system shows hysteresis from loading and unloading of its elastic elements, whereas Levio in \textit{active assistance} mode shows a positive work loop. The passive system's lower assistance early in bending is likely due to initial slack between the user and the chest harness, which must be taken up before the load-cell registers interaction force. We did not observe this in our previous study~\cite{refai2024benchmarking}, which estimated assistance indirectly from marker trajectories and the apparent spring stiffness at each pose. This highlights the value of integrated sensing for quantifying actual assistance and interface losses in passive systems.

Fig.~\ref{fig:active-passive-difference} shows the participant-wise difference in assistance between the \textit{active assistance} and \textit{passive} modes. Levio provided less assistance than the passive system during bending, and therefore less resistance. During lifting it provided higher, and more sustained, assistance, with the largest difference at about 35°--50°. The controller thus shifted assistance toward the lifting phase rather than increasing it uniformly. The larger between-participant spread (shaded SD) during lifting may reflect variability in loading, fit, and potential user-interface. Trends beyond $\sim$70° should be interpreted cautiously, as not all participants bent that far.

Levio in \textit{active assistance} mode also received favourable usability and discomfort scores (Tables~\ref{tab:sus-results}, \ref{tab:global-vas-results}, \ref{tab:body-region-vas-results}). Lower-back discomfort was slightly higher in \textit{zero torque} mode (Table~\ref{tab:body-region-vas-results}), possibly because the device weight then rests on the hip and back. This suggests the device may be more fatiguing to wear when un-assisted. With assistance enabled, discomfort was lower, possibly because the added interaction forces stabilise the device on the user. 

There are a few limitations to the study. Stoop assistance was successfully demonstrated, but squat trials revealed thigh-pad migration and a mismatch between the measured torso--pelvis angle and the effective exoskeleton configuration. Including an IMU on the thigh could be beneficial to measure bending during squat or other lifting tasks, but as it needed an additional sensing element, it was not implemented in this study. Moreover, joint encoder information could also be used to estimate exoskeleton/human pose.

The retrofit also increased the total mass from 4.3 to approximately 7 kg. Optimizing the power supply and electronics could reduce some of this weight.  The deformation of the thigh connector prevented verification of the original 40 Nm per-side design target, so participant testing was limited to 25 Nm per side. Improving the interface design could support higher assistive torques. Finally, the small sample and descriptive analysis preclude conclusions about statistical significance.

Future work should prioritise connector redesign and structural verification, mass reduction, and improved sensing of the effective exoskeleton angle. Drivetrain friction, compliance, hysteresis, and torque-response dynamics should also be identified before evaluating more advanced controllers and their biomechanical effects.

\begin{table}[!t]
\centering
\caption{Body-region discomfort VAS for stoop (mean $\pm$ SD, $n=9$).}
\label{tab:body-region-vas-results}
\small
\setlength{\tabcolsep}{2.5pt}
\begin{tabular}{lccc}
\hline
\textbf{Region} & \textbf{Passive} & \textbf{Zero-torque} & \textbf{Active} \\
\hline
Chest & $1.7\pm1.5$ & $1.4\pm1.2$ & $1.2\pm0.8$ \\
Upper back & $1.4\pm0.9$ & $1.8\pm1.6$ & $1.0\pm0.7$ \\
Lower back & $2.8\pm1.9$ & $3.3\pm2.9$ & $1.4\pm1.0$ \\
Abdomen & $1.6\pm1.7$ & $2.1\pm2.5$ & $1.3\pm1.4$ \\
Front legs & $2.8\pm1.9$ & $2.3\pm1.7$ & $2.1\pm1.4$ \\
Back legs & $1.1\pm0.9$ & $1.4\pm1.6$ & $0.9\pm0.8$ \\
\hline
\end{tabular}
\end{table}

\section{Conclusions}
This work presents Levio, an active retrofit of the passive Laevo FLEX V3.0 using commercial actuators, sensing, power supply, and electronics. This demonstrates the feasibility of retrofitting a passive back-support exoskeleton while preserving its wearable architecture and passive alignment degrees of freedom. The retrofit increased the total mass from \SI{4.3}{kg} to approximately \SI{7}{kg}, and provided a maximum assistance of \SI{25}{Nm} per side. Within this operating envelope, Levio provided programmable assistance during stoop lifting. Compared with the passive system, it left lifting kinematics unchanged, shifted peak assistance from the fully bent pose to the lifting phase, and received favourable usability and discomfort scores. Levio thus provides a functional platform for evaluating alternative control strategies. The study demonstrates that retrofit solutions could be a potential commercial add-on for existing passive systems.

\bibliographystyle{IEEEtran} 
\bibliography{references.bib}

\vspace{12pt}

\end{document}